\documentclass[runningheads]{llncs}
\usepackage{graphicx}
\usepackage{amsmath}
\usepackage{tabularx} 
\usepackage{booktabs} 
\usepackage{siunitx} 
\usepackage{multirow}
\usepackage{multicol}
\usepackage{subfig}
\usepackage{verbatim}
\usepackage{color}

\usepackage{hyperref}

\newcommand{\paul}[1]{{\color{black}#1}}
\newcommand{\af}[1]{{\color{black}#1}}

\begin{document}
%
\title{Offline Signature Verification by Combining Graph Edit Distance and Triplet Networks\vspace{-1mm}}
%
\titlerunning{Combining Graph Edit Distance and Triplet Networks}
%

\author{Paul Maergner\inst{1} \and
Vinaychandran~Pondenkandath\inst{1} \and
Michele~Alberti\inst{1} \and
Marcus~Liwicki\inst{1} \and
Kaspar Riesen\inst{2} \and
Rolf Ingold\inst{1} \and
Andreas Fischer\inst{1,3}}
\authorrunning{Maergner et al.}
%
\institute{\af{University of Fribourg, DIVA group, 1700 Fribourg, Switzerland, \\ Email: \email{\{firstname\}.\{lastname\}@unifr.ch}} \and
\af{University of Applied Sciences and Arts Northwestern Switzerland, Institute for Information Systems, 4600 Olten, Switzerland, Email: \email{kaspar.riesen@fhnw.ch}}  \and
\af{University of Applied Sciences and Arts Western Switzerland, Institute of\\Complex Systems, 1700 Fribourg, Switzerland, Email: \email{andreas.fischer@hefr.ch}}
}

\maketitle              

\vspace{-3mm}
\begin{abstract}
\af{Biometric authentication by means of handwritten signatures is a challenging pattern recognition task, which aims to infer a writer model from only a handful of genuine signatures. In order to make it more difficult for a forger to attack the verification system, a promising strategy is to combine different writer models. In this work, we propose to complement a recent structural approach to offline signature verification based on graph edit distance with a statistical approach based on metric learning with deep neural networks. On the MCYT and GPDS benchmark datasets, we demonstrate that combining the structural and statistical models leads to significant improvements in performance, profiting from their complementary properties.}


\vspace{-2mm}
\end{abstract}
%
%
%
\section{Introduction}
\af{To this day, handwritten signatures have remained a widely used and accepted means of biometric authentication. Automatic signature verification is an active field of research, accordingly, and the current state of the art achieves levels of accuracy similar to that of other biometric verification systems~\paul{\cite{Hafemann2017review,Impedovo2008}}.}
Usually, two cases of signature verification are differentiated: the \emph{offline} case, where only a static image of the signature is available, and the \emph{online} case, where \af{additional} dynamic information like \af{the velocity} is available. 
Due to the lack of this information, offline signature verification applies to more use cases, but it is also considered the more challenging task.

Most state-of-the-art approaches to offline signature verification rely on statistical pattern recognition, i.e. signatures are represented using fixed-size feature vectors. 
These vector representations are often generated using handcrafted feature extractors leveraging either \emph{local information}, such as local binary patterns, histogram of oriented gradients, or Gaussian grid features taken from signature contours \cite{Yilmaz2011OfflineFeatures}, or \emph{global information}, e.g. geometrical features like Fourier descriptors, number of branches in the skeleton, number of holes, moments, projections, distributions, position of barycenter, tortuosities, directions, curvatures and chain codes \cite{Impedovo2008,Plamondon1989}. 
\af{More recently, with the advent of deep learning, we observe} a shift away from handcrafted features towards learning features directly from the images using deep convolutional neural networks (CNN)~\cite{Hafemann2017}.  

Another way of approaching signature verification is by using graphs and structural pattern recognition. 
Graphs offer a more powerful representation formalism that \af{can} be beneficial for signature verification. 
For example, by capturing local information in nodes and their relations in the global structure using edges. 
But the representational power of graphs comes at the price of high computational complexity. 
This is probably why graphs have \af{only been used rather rarely} for signature verification in the past. 
\af{Examples include the work of} Sabourin et al.~\cite{sabourin94structural} (signatures represented based on stroke primitives), Bansal et al.~\cite{Bansal2009} (modular graph matching approach), and Fotak et al.~\cite{Fotak2011} (basic concepts of graph theory). \af{More recently}, a structural approach for signature verification has been introduced by Maergner et al.~\cite{maergner2017icdar}. 
They propose a general signature verification framework based on the graph edit distance between labeled graphs. 
They \af{employ a} bipartite approximation framework~\cite{Riesen2009} to reduce the computational complexity and \af{report} promising verification results using so-called keypoint graphs.

\begin{figure}[t]
\centering
\includegraphics[width=.80\columnwidth,trim={0mm 0mm 0mm 0mm},clip]{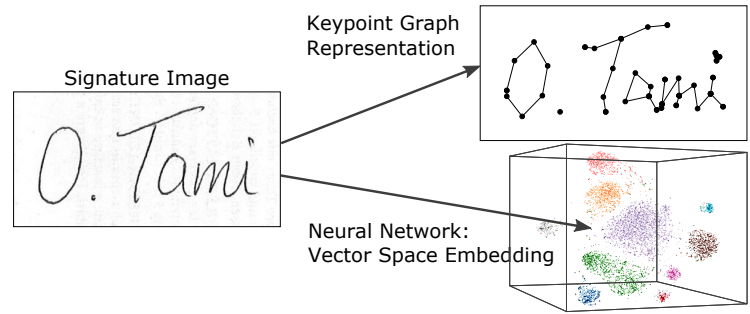}
\caption{Proposed structural and statistical signature image representations \label{fig:example}}
\end{figure}

\af{In this paper, we argue that structural and statistical signature models are quite different, with complementary strengths, and thus well-suited for multiple classifier systems. As illustrated in Fig.~\ref{fig:example}, we propose to combine the graph-based approach of Maergner et al.~\cite{maergner2017icdar} with a statistical model inspired by recent advances in the field of deep learning, namely metric learning by means of a deep CNN~\cite{he2016deep} with the triplet loss function~\cite{hoffer2015}. Such deep triplet networks can be used to embed signature images in a vector space, where signatures of the same user have a small distance and signatures of different users have a large distance.} To our knowledge, this is the first combination of a graph-based approach and a deep neural network based approach for the task of signature verification.

\af{In the remainder, the structural approach is described in Section~\ref{sec:graph}, the statistical approach in Section~\ref{sec:nn}, and the proposed combined system in Section~\ref{sec:sigver}. Afterwards, we present our experimental results in Section~\ref{sec:exp} and draw conclusions in Section~\ref{sec:concl}.}



\section{Structural Graph-Based Approach \label{sec:graph}}
The structural approach used in this paper has been proposed by Maergner et al. in~\cite{maergner2017icdar}. 
Two signature images are compared by first binarizing and skeletonizing the image, then creating keypoint graphs from each skeleton image, and lastly comparing the two graphs using an approximation of the graph edit distance. 
In the following subsections, we briefly review these steps. 
For a more detailed description, see~\cite{maergner2017icdar}.


\subsection{Keypoint Graphs}
Formally, a labeled graph is defined as a four-tuple \mbox{$g = (V,E,\mu,\nu)$}, where $V$ is the finite set of nodes, $E\subseteq V \times V$ is the set of edges, $\mu:V\rightarrow L_V$ is the node labeling function, \af{and} $\nu: E \rightarrow L_E$ is the edge labeling function. 

Keypoint graphs are created from points extracted from the skeleton image. 
Specifically, the nodes in the graph stand for certain points on the skeleton and are labeled with their coordinates. 
These points are end- and junction-points of the skeleton as well as additional points sampled in equidistant intervals of $D$.
Unlabeled and undirected edges connect the nodes that are \af{connected} on the skeleton.
The node labels are centered so that their average is $(0, 0)$. 
See Fig.~\ref{fig:example} for an example of a keypoint graph.

\subsection{Graph Edit Distance}
Graph edit distance (GED) offers a way to compare any kind of labeled graph given an appropriate cost function. 
This makes GED one of the most flexible graph matching approaches. 
It calculates the cost of the \af{lowest-cost} edit path that transforms graph $g_1=(V_1,E_1,\mu_1,\nu_1)$ into graph $g_2=(V_2,E_2,\mu_2,\nu_2)$. 
An edit path is a sequence of edit operations, \af{for each of which a certain cost is defined}. 
Commonly, substitutions, deletions, \af{and} insertions of nodes and edges \af{are considered as edit operations}. 
The main disadvantage of GED is its computational complexity since it is exponential in the number of nodes in the two graphs, $O(|V_1|^{|V_2|})$.

This issue can be addressed by using an approximation of GED. 
In this paper, the bipartite approximation framework proposed by Riesen and Bunke~\cite{Riesen2009} is applied. 
The computation of GED is reduced to an instance of a \emph{linear sum assignment problem} with cubic complexity, $O\big((V_1+V_2)^3\big)$. 
\af{For signature verification,} the lower bound introduced in~\cite{Riesen2014} is considered.

The cost function is defined in the following way. 
The cost of a node substitution is the Euclidean distance between the node labels.
For node deletion and insertion, a constant cost $C_\text{node}$ is used.
For edges, the substitution cost is set to zero.
The edge deletion and insertion cost is set to a constant value $C_\text{edge}$.


\af{Finally, the} graph edit distance is normalized by dividing by the \af{maximum} graph edit distance, viz. the cost of deleting all nodes and edges from the first graph and inserting all the nodes and edges of the second graph. 
Thus, the graph-based dissimilarity is in $[0, 1]$ and describes how \af{large} the graph edit distance is \af{when compared with} the maximum graph edit distance. 
Formally, the graph-based dissimilarity of two signature images is defined as follows:
\begin{equation}
d_{\text{GED}}(r,t) = \frac{\text{GED}(g_r,g_t)}{\text{GED}_{\text{max}}(g_r,g_t)} \text{,}
\end{equation}
where 
$g_r$ and $g_t$ are the keypoint graphs of the signatures images $r$ and $t$ respectively, 
$\text{GED}(g_r,g_t)$ is the lower bound of the graph edit distance between $g_r$ and $g_t$, 
and $\text{GED}_{\text{max}}(g_r,g_t)$ is the \af{maximum} graph edit distance between $g_r$ and $g_t$.

\section{\af{Statistical} Neural Network\af{-Based} Approach}\label{sec:nn}

We train a \af{deep CNN~\cite{he2016deep}} using a triplet-based learning \af{method} to embed images of signatures into a high-dimensional space where the distance of two signatures reflect their similarity, i.e. two signatures of the same \af{user} are close together and signatures from different \af{users} are far apart. 
An \af{exemplary} visualization of the vectors produced by such model \af{is shown in Fig.~\ref{fig:example},} where points of the same class are grouped together in clusters.
\af{This approach has been investigated in the recent past for several image matching problems with promising success, including~\cite{balntas2016,hoffer2015,zagoruyko2015}.}

\subsection{Triplet-Based \af{Learning}}
\label{sec:neural}

A triplet is a tuple of three signatures $\{a, p, n\}$ where $a$ is the anchor (reference signature), $p$ is the positive sample (a signature from the same \af{user}) and $n$ is the negative sample (a signature from another \af{user}).
The neural network is then trained to minimize the loss function defined as:
\begin{equation}\label{eq:loss}
    L(\delta_+,\delta_-) = max(\delta_+ - \delta_- + \mu, 0)\text{\af{,}}
\end{equation}
\noindent\af{where} $\delta_+$ and $\delta_-$ are the Euclidean distance between anchor-positive and anchor-negative pairs in the feature space and $\mu$ is the margin used.

\subsection{\af{Signature Image Matching}}
\label{sec:neural_distance}
We define the neural network as the function $f$ \af{that} embeds a signature image into a latent space as previously described.
The \af{dissimilarity} of two signature images $r$ and $t$ can now be defined as the Euclidean distance of their embedding vectors. \af{Formally,}
\begin{equation}
    d_{\text{neural}}(r, t) = \lVert f(r) - f(t) \rVert_2\text{\af{.}}
\end{equation}

\section{\af{Combined} Signature Verification System}\label{sec:sigver}
A signature verification system has to decide whether an unseen signature image is a genuine signature of the claimed user. 
This decision is being made by calculating a dissimilarity score  between the \emph{reference} signature of the claimed user and the unseen signature.
The signature is accepted if this dissimilarity score (see Eq.~\ref{eq:score1}~or~\ref{eq:score2}) is below a certain threshold, otherwise the signature is rejected.

\subsection{User-based Normalization}
It is expected that the users have different \af{intra}-user variability.
Therefore, each dissimilarity score is normalized using the average dissimilarity score between the reference signatures of the current user as suggested in~\cite{maergner2017icdar}. 
Formally, 
\begin{equation}
\hat{d}(r, t) =  \frac{d(r,t)}{\delta(R)} \text{,}
\end{equation}
where $t$ is a questioned signature image, $r \in R$ is a reference signature image, $R$ is the set of all reference signature images of the current users, and
\begin{equation*}
\delta(R) = \frac{1}{|R|}\sum_{r \in R}\min_{s \in R \setminus {r}}d(r, s) \text{.}
\end{equation*}

\subsection{Signature Verification Score}
The minimum dissimilarity over all reference signatures R of the claimed user to the questioned signature t is used as signature verification score. Formally, 
\begin{equation}
\label{eq:score1}
d(R,t) = \min_{r \in R} \hat{d}(r,t)
\end{equation}

\subsection{Multiple Classifier System \label{sec:mcs}}
We propose a multiple classifier system (MCS) as a linear combination of the graph-based dissimilarity and the neural network based dissimilarity. 
Z-score normalization based on all reference signature images in the current data set is applied to each dissimilarity score before the combination. 
Formally, we define
\begin{equation}
\label{eq:score2}
\af{d_{\text{MCS}}(R,t) = \min_{r \in R} \Big(\hat{d}^*_{\text{GED}}(r,t) + \hat{d}^*_{\text{neural}}(r,t) \Big) \text{,}}
\end{equation}
where $\hat{d}^*$ is the z-score normalized dissimilarity score.

\section{Experimental Evaluation}\label{sec:exp}
We evaluate the performance on two publicly available benchmark data sets by measuring \af{the \emph{equal error rate} (EER).} 
The EER is the point where the false acceptance rate and the false rejection rate are equal in the \emph{detection error tradeoff (DET)} curve.
Two kinds of forgeries are tested: 
\emph{skilled forgeries} (SF), which are forgeries created with information about the user's signature, 
and so-called \emph{random forgeries}\paul{\footnote{\paul{This term is mainly used in the pattern recognition community and it might be confusing for readers from other fields. For more details, see~\cite{Malik2012a}.}}}
(RF), which are genuine signatures of other users that are used in a brute force attack.

\subsection{Data Sets}
In our evaluation, we use the following publicly available signature data sets:

\begin{itemize}
    \item \textbf{GPDSsynthetic-Offline:}
            Ferrer et al. introduced this data set in~\cite{Ferrer2015}. 
            It contains 24 genuine signatures and 30 skilled forgeries for $4,000$ synthetic users. 
            This data set replaces previous signatures databases from the GPDS group, which are not available anymore.
            
            We use four subsets of this data set: one containing the first 75 users, and three containing the last 10, 100, or 1000 users. 
            These subsets are called \emph{GPDS-75}, \emph{GPDS-last10}, \emph{GPDS-last100}, and \emph{GPDS-last1000} respectively. 
    
    \item \textbf{MCYT-75:}
            This data set is part of the MCYT baseline corpus introduced by Ortega-Garcia et al. in\af{~\cite{Fierrez-Aguilar2004,Ortega-Garcia2003}}. 
            It contains 75 users with 15~genuine signatures and 15~skilled forgeries each. 
\end{itemize}

\subsection{Tasks}
We distinguish two tasks depending on the number of references available for each user. 
Five genuine signatures per user (\emph{R5}) or ten genuine signatures per user (\emph{R10}). 
In both cases, the remaining genuine signatures are used for testing in both the skilled forgery (SF) and in the random forgery (RF) evaluation. 
The SF evaluation is performed using all available skilled forgeries for each user. 
The RF evaluation is carried out using the first genuine signature of all other users in the data set as random forgeries.
For example for the GPDS-75 R10 tasks, that gives us $75 \cdot 10 = 750$ reference signatures, $75 \times 14 = 1,050$ genuine signatures, $75 \times 30 = 2,250$ skilled forgeries, and $75 \times 74 = 5,550$ random forgeries.

\subsection{Setup}
\subsubsection{Graph Parameter Validation}
For the keypoint graph \af{extraction}, we \af{use} $D=25$, which has been proposed in \cite{maergner2017icdar}. The cost function \af{parameters} $C_\text{node}$ and $C_\text{edge}$ \af{are} validated on the GPDS-last100 data set using the random forgery evaluation. No skilled forgeries \af{are} used. We \af{perform} a grid search over $C_\text{node} \in \{10, 15, \dots, 60\}$ and $C_\text{edge} \in \{10, 15, \dots, 60\}$. The best results have been achieved using $C_\text{node} = 25$ and $C_\text{edge} = 45$. We use these parameters \af{in} our experiments on GPDS-75 and MCYT-75.

\subsubsection{Neural Network Training}
We use the ResNet18 architecture~\cite{he2016deep}, which is an $18$ layer deep variant of a convolutional neural network that uses shortcut connections between layers to tackle the vanishing gradient problem. 

We train three different models using the DeepDIVA\paul{\footnote{\paul{\url{https://github.com/DIVA-DIA/DeepDIVA} (April 29, 2018)}}} 
framework~\cite{alberti2018deepdiva} for the task of embedding the signature images in the vector space, where each of the models differs with respect to how much data is used for training (GPDS-last10, GPDS-last100, or GPDS-last1000). 
We call these systems NN-last10, NN-last100, and NN-last1000 respectively.
For each person in the data set, there are 24 genuine images. We use 16 of them for training and the remaining 8 for validating the performance of the model. Skilled forgeries \af{are not} used for training.

The network is trained using the Stochastic Gradient Descent (SGD) optimizer with a learning rate of $0.01$ and momentum of $0.9$.  

\subsection{Results on MCYT-75 and GPDS-75}
\begin{figure*}[t]
\centering
\vspace{-4mm}
\subfloat[Skilled Forgeries \label{fig:det-curves-sf-gpds-r10}]{
    \includegraphics[width=.38\textwidth,trim={24mm 0mm 30mm 10mm},clip]{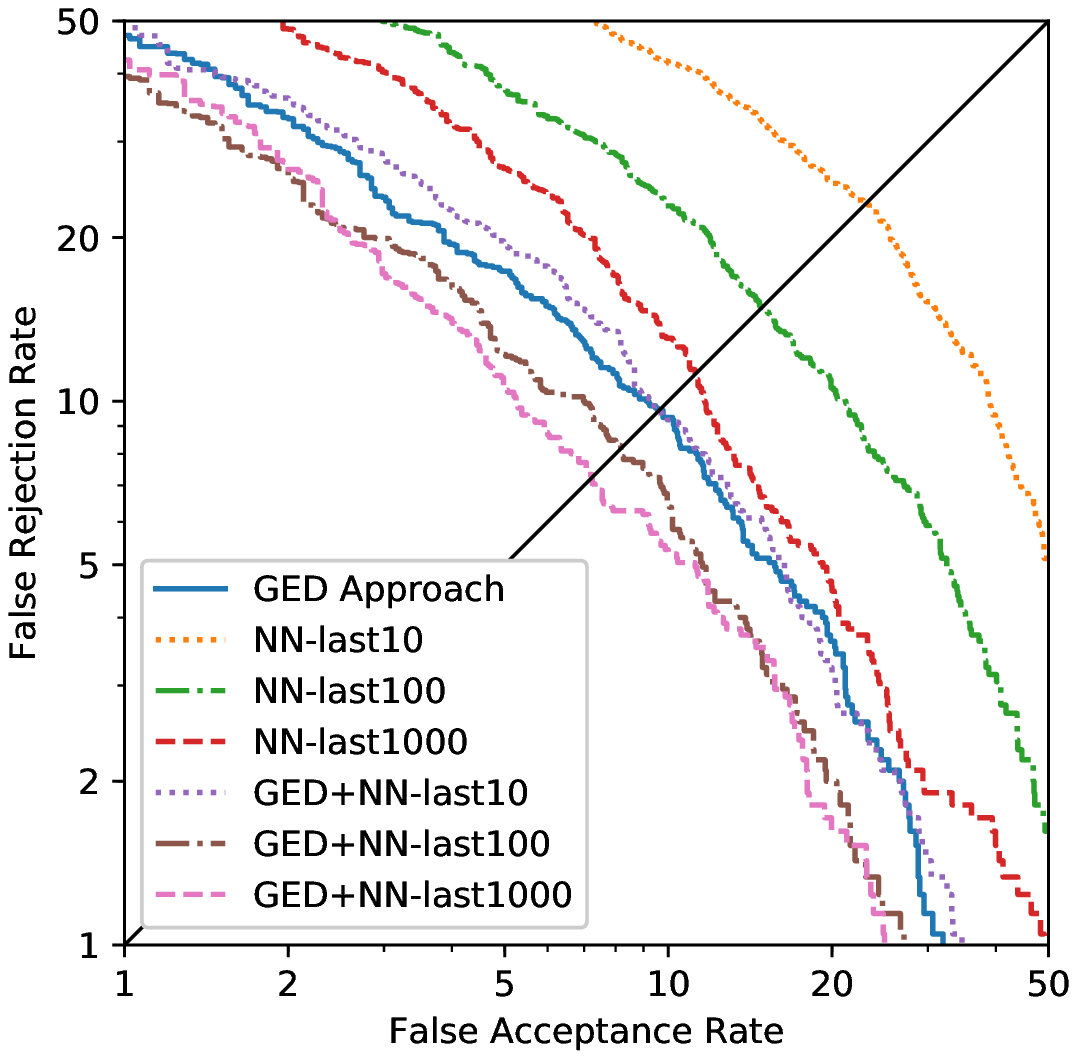}
}
\subfloat[Random Forgeries \label{fig:det-curves-rf1-gpds-r10}]{
    \includegraphics[width=.38\textwidth,trim={24mm 0mm 30mm 10mm},clip]{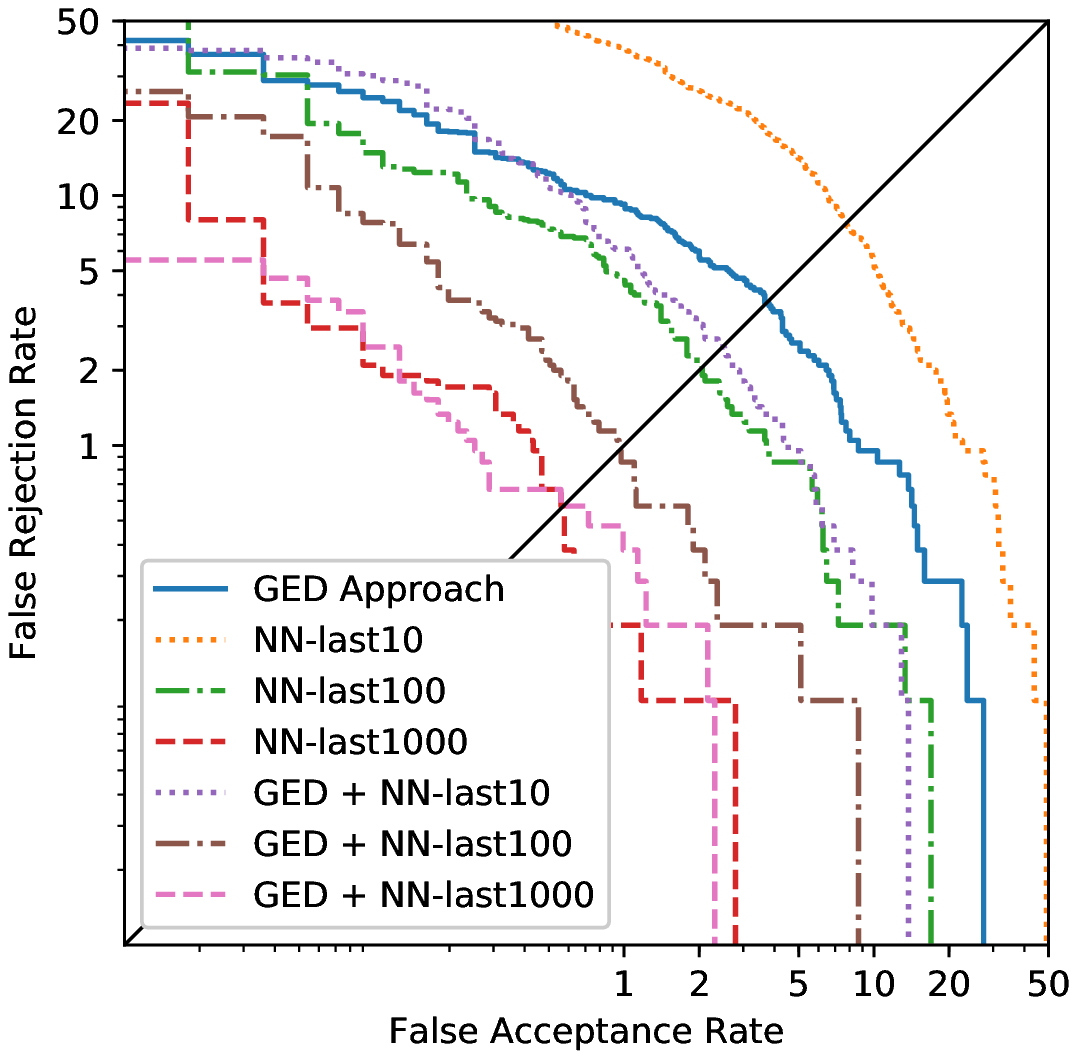}
}
\caption{DET curves for GPDS-75 R10 \label{fig:det-curves-sf-rf}}
\vspace{-4mm}
\end{figure*}

The EER results on GPDS-75 and MCYT-75 for both RF and SF are shown in Table~\ref{tab:results}. 
In all but one case, the combination of the GED approach and the neural network achieves better results than the \af{best individual system}.
The neural networks trained on GPDS-last100 and GPDS-last1000 are on its own significantly better on the RF task.
We can see that NN-last1000 is more specialized on the RF task on the GPDS-75 data set while losing performance on the MCYT-75 data set.
Two DET curves are shown in \af{Fig.~\ref{fig:det-curves-sf-rf}.}

\subsection{Comparison with State-of-the-Art}
Many different evaluation protocols are used for signature verification. 
To allow a fair comparison, we have to follow the same protocol. 
In the following, we present EER results using two different protocols and compare our results with other published results.

\begin{table}
\setlength{\tabcolsep}{5pt}
    \caption{\textbf{EER on GPDS-75/MCYT-75}. Results on skilled forgeries (SF) and on random forgeries (RF) using the first 5 or 10 genuine as references (R5/R10).}
    \centering
    \label{tab:results}
    \begin{tabularx}{\textwidth}{X
    S[table-format=2.2, input-decimal-markers = .]
    S[table-format=2.2, input-decimal-markers = .]
    S[table-format=2.2, input-decimal-markers = .]
    S[table-format=2.2, input-decimal-markers = .]
    S[table-format=2.2, input-decimal-markers = .]
    S[table-format=2.2, input-decimal-markers = .]
    S[table-format=2.2, input-decimal-markers = .]
    S[table-format=2.2, input-decimal-markers = .]
    }
    \toprule
     & \multicolumn{4}{c}{\textbf{{GPDS-75}}} & \multicolumn{4}{c}{\textbf{{MCYT-75}}} \\
    \cmidrule(lr){2-5}
    \cmidrule(lr){6-9}
    \textbf{System}             & \multicolumn{2}{S}{\textbf{RF}} & \multicolumn{2}{S}{\textbf{SF}} & \multicolumn{2}{S}{\textbf{RF}} & \multicolumn{2}{S}{\textbf{SF}}\\
    \cmidrule(lr){2-3}
    \cmidrule(lr){4-5} 
    \cmidrule(lr){6-7} 
    \cmidrule(lr){8-9}
                 & \textbf{R5} & \textbf{R10} & \textbf{R5} & \textbf{R10} & \textbf{R5} & \textbf{R10} & \textbf{R5} & \textbf{R10} \\
    \midrule
    {GED approach}      &  4.90 &  3.71 & 11.69 &  9.60    &  5.86 &  2.65 & 20.09 & 13.60 \\[1ex]
    {NN-last10}         & 10.40 &  7.71 & 25.87 & 23.11    &  6.47 &  4.79 & 19.56 & 17.16 \\
    {GED + NN-last10}   &  4.00 &  2.47 & 12.04 &  9.51    &  3.19 &  1.59 & 16.53 & 11.29 \\[1ex]
    {NN-last100}        &  3.28 &  2.05 & 17.96 & 14.84    &  3.59 &  1.59 & 20.36 & 12.80 \\
    {GED + NN-last100}  &  2.16 &  0.95 &  9.82 &  8.18    &  \multicolumn{1}{r}{\textbf{2.79}} &  1.41 &  \multicolumn{1}{r}{\textbf{15.56}} &  \multicolumn{1}{r}{\textbf{10.40}} \\[1ex]
    {NN-last1000}       &  0.68 &  \multicolumn{1}{r}{\textbf{0.56}} & 13.29 & 11.20    &  3.73 &  1.15 & 19.02 & 13.78 \\
    {GED + NN-last1000} & \multicolumn{1}{r}{\textbf{0.65}}  &  \multicolumn{1}{r}{\textbf{0.56}} &  \multicolumn{1}{r}{\textbf{9.24}} &  \multicolumn{1}{r}{\textbf{7.24}}  &  2.92 &  \multicolumn{1}{r}{\textbf{0.79}} & 17.69 & 11.11 \\[1ex]
    \bottomrule
    \end{tabularx}
\end{table}

\begin{table}
    \caption{\textbf{Comparison on GPDS-75/MCYT-75}.
    Average EER results over 10 random selections of ten reference signatures. Evaluated on GPDS-75 and MCYT-75 for random forgeries (RF) and skilled forgeries (SF).}
    \centering
    \label{tab:results-gpds}
    \begin{tabularx}{\columnwidth}{X    
    S[table-format=2.2,detect-weight]
    S[table-format=2.2,detect-weight]
    S[table-format=2.2,detect-weight]
    S[table-format=2.2,detect-weight]
    S[table-format=2.2,detect-weight]
    }
    \toprule
    \multirow{2}{*}{\bfseries System} & \multicolumn{2}{c}{\bfseries GPDS-75 R10} & \multicolumn{1}{c}{\hphantom{space}} & \multicolumn{2}{c}{\bfseries MCYT-75 R10} \\
    \cmidrule(lr){2-3} \cmidrule(lr){5-6}
                 & \multicolumn{1}{l}{\bfseries RF} & \multicolumn{1}{r}{\bfseries SF} & \multicolumn{1}{c}{ } & \multicolumn{1}{l}{\bfseries RF} & \multicolumn{1}{r}{\bfseries SF} \\
    \midrule
    Ferrer et al.~\cite{Ferrer2012}$^7$         &  \multicolumn{1}{l}{0.76*} & \multicolumn{1}{r}{16.01} & & \multicolumn{1}{l}{\textbf{0.35}*} & \multicolumn{1}{r}{11.54} \\
    Maergner et al.~\cite{maergner2017icdar}    &  \multicolumn{1}{l}{2.73}  & \multicolumn{1}{r}{8.29}  & & \multicolumn{1}{l}{2.83}  & \multicolumn{1}{r}{12.01} \\ [1ex]
    Proposed GED approach                 		&  \multicolumn{1}{l}{2.75}  & \multicolumn{1}{r}{8.31}  & & \multicolumn{1}{l}{2.67}  & \multicolumn{1}{r}{11.42} \\
    Proposed NN-last1000                        &  \multicolumn{1}{l}{0.44}  & \multicolumn{1}{r}{10.79} & & \multicolumn{1}{l}{1.57}  & \multicolumn{1}{r}{12.24} \\
    Proposed GED + NN-last1000                  &  \multicolumn{1}{l}{\textbf{0.41}}  &  \multicolumn{1}{r}{\textbf{6.49}} & \multicolumn{1}{c}{ } &  \multicolumn{1}{l}{1.05}  &  \multicolumn{1}{r}{\textbf{9.15}} \\
    \bottomrule
    \multicolumn{5}{l}{*: All genuine signatures of other users as RF} \\
    \end{tabularx}
\end{table}

\begin{table}
    \caption{\textbf{Comparison on MCYT-75 R5/R10}. EER results for skilled forgeries (SF) and random forgeries (RF) using an \emph{a posteriori} user-dependent score normalization. The first 5 or 10 genuine signatures are used as references for R5 and R10 respectively.}
    \centering
    \label{tab:results-mcyt}
    \begin{tabularx}{\columnwidth}{X    
    S[table-format=2.2,detect-weight]
    S[table-format=2.2,detect-weight]
    S[table-format=2.2,detect-weight]
    S[table-format=2.2,detect-weight]
    S[table-format=2.2,detect-weight]
    }
    \toprule
    \multirow{2}{*}{\bfseries System} & \multicolumn{2}{c}{\bfseries MCYT-75 R5} & \multicolumn{1}{c}{\hphantom{space}} & \multicolumn{2}{c}{\bfseries MCYT-75 R10} \\
    \cmidrule(lr){2-3} \cmidrule(lr){5-6}
                 & \multicolumn{1}{l}{\bfseries RF} & \multicolumn{1}{r}{\bfseries SF} & \multicolumn{1}{c}{ } & \multicolumn{1}{l}{\bfseries RF} & \multicolumn{1}{r}{\bfseries SF} \\
    \midrule
    Alonso-Fernandez et al.~\cite{Alonso-Fernandez2007AutomaticSignature}   & \multicolumn{1}{l}{9.79*}  & \multicolumn{1}{r}{23.78} & & \multicolumn{1}{l}{7.26*}  & \multicolumn{1}{r}{22.13} \\
    Fierrez-Aguilar et al.~\cite{Fierrez-Aguilar2004}                       & \multicolumn{1}{l}{2.69**} & \multicolumn{1}{r}{11.00} & & \multicolumn{1}{l}{1.14**} & \multicolumn{1}{r}{9.28}  \\
    Gilperez et al.~\cite{Gilperez2008Off-lineFeatures}                     & \multicolumn{1}{l}{2.18*}  & \multicolumn{1}{r}{\textbf{10.18}} & & \multicolumn{1}{l}{1.18*}  & \multicolumn{1}{r}{\textbf{6.44}}  \\
    Maergner et al.~\cite{maergner2017icdar}                                & \multicolumn{1}{l}{2.40}   & \multicolumn{1}{r}{14.49} & & \multicolumn{1}{l}{1.89}   & \multicolumn{1}{r}{11.64} \\[1ex]
    Proposed GED approach                                                   & \multicolumn{1}{l}{2.45}   & \multicolumn{1}{r}{14.84} & & \multicolumn{1}{l}{1.89}   & \multicolumn{1}{r}{12.27} \\ 
    Proposed NN-last100                                                     & \multicolumn{1}{l}{2.14}   & \multicolumn{1}{r}{15.02} & & \multicolumn{1}{l}{1.77}   & \multicolumn{1}{r}{13.16} \\
    Proposed GED + NN-last100                                               & \multicolumn{1}{l}{\textbf{0.92}}   & \multicolumn{1}{r}{10.67} & & \multicolumn{1}{l}{\textbf{0.25}}   & \multicolumn{1}{r}{10.13} \\
    \bottomrule
    \multicolumn{5}{l}{ *: All genuine signatures of other users as RF} \\
    \multicolumn{5}{l}{**: First 5 genuine signatures from each other user as RF.} \\
    \end{tabularx}
\end{table}

\subsubsection{Comparison on GPDS-75 and MCYT-75}
This evaluation is performed by selecting 10 reference signatures randomly\footnote{We use the same random selections for all our results.}
and average the results over 10 runs. 
Table~\ref{tab:results-gpds} shows our results using the same protocol compared with the previously published results: 
results published in~\cite{maergner2017icdar} and results presented on the GPDS website\footnote{\label{footnote:gpds}\url{http://www.gpds.ulpgc.es/downloadnew/download.htm} (\af{April 29}, 2018)}, which have been achieved using the system published in~\cite{Ferrer2012}. 
The proposed combination of \af{the} GED approach and NN-last1000 achieves the lowest EER in all tasks except for random forgeries on MCYT-75.

\subsubsection{Comparison on MCYT-75}
A group of publications has presented results on the MCYT-75 data set using the a posteriori user-depended score normalization introduced in~\cite{Fierrez-Aguilar2004TargetVerification}.
By applying this normalization, all user scores are aligned so that the EER threshold is the same for all users. 
Table~\ref{tab:results-mcyt} shows the published results as well as our results using the same normalization. The combination of GED and NN-last100 achieves \af{results in the middle ranks for the SF task and the overall best results for the RF task.}

\section{Conclusions and Outlook}\label{sec:concl}
\af{Combining structural and statistical models has significantly improved the signature verification performance on the MCYT-75 and GPDSsynthetic-Offline benchmark datasets. The structural model based on approximate graph edit distance achieved better results against skilled forgeries, while the statistical model based on metric learning with deep triplet networks achieved better results against a brute-force attack with random forgeries. The proposed system was able to combine these complementary strengths and has proven to generalize well to unseen users, which have not been used for model training and hyperparameter optimization.}

\af{We can see several lines of future research. For the structural method, more graph-based representations and cost functions may be explored in the context of graph edit distance. For the statistical method, synthetic data augmentation may lead to a more accurate vector space embedding.} Finally, we believe that there is a \af{great} potential in combining even more structural and statistical classifiers \af{into} one large multiple classifier system. Such a system \af{is expected to further} improve the robustness of biometric authentication.

\section*{Acknowledgment}
This work has been supported by the Swiss National Science Foundation project 200021\_162852. \\
The final authenticated publication is available online at \\ \url{https://doi.org/10.1007/978-3-319-97785-0\_45}.

%
%
%
\bibliographystyle{splncs04}
\bibliography{mendeley}

\begin{thebibliography}{10}
\providecommand{\url}[1]{\texttt{#1}}
\providecommand{\urlprefix}{URL }
\providecommand{\doi}[1]{https://doi.org/#1}

\bibitem{alberti2018deepdiva}
Alberti, M., Pondenkandath, V., W\"ursch, M., Ingold, R., Liwicki, M.:
  {DeepDIVA: A Highly-Functional Python Framework for Reproducible
  Experiments}. {Submitted at Int. Conference on Frontiers in Handwriting
  Recognition}  (2018)

\bibitem{Alonso-Fernandez2007AutomaticSignature}
Alonso-Fernandez, F., Fairhurst, M., Fierrez, J., Ortega-Garcia, J.: Automatic
  measures for predicting performance in off-line signature. In: {Proc. 14th
  Int. Conf. on Image Processing}. pp. 369--372 (2007)

\bibitem{balntas2016}
Balntas, V., Riba, E., Ponsa, D., Mikolajczyk, K.: Learning local feature
  descriptors with triplets and shallow convolutional neural networks. In:
  Proc. of the British Machine Vision Conference (BMVC) (Sept 2016)

\bibitem{Bansal2009}
Bansal, A., Gupta, B., Khandelwal, G., Chakraverty, S.: Offline signature
  verification using critical region matching. Int. Journal of Signal
  Processing, Image Processing and Pattern Recognition  \textbf{2}(1),  57--70
  (2009)

\bibitem{Ferrer2015}
Ferrer, M.A., Diaz-Cabrera, M., Morales, A.: {Static Signature Synthesis: A
  Neuromotor Inspired Approach for Biometrics}. IEEE Transactions on Pattern
  Analysis and Machine Intelligence  \textbf{37}(3),  667--680 (mar 2015)

\bibitem{Ferrer2012}
Ferrer, M.A., Vargas, J.F., Morales, A., Ordonez, A.: {Robustness of Offline
  Signature Verification Based on Gray Level Features}. IEEE Transactions on
  Information Forensics and Security  \textbf{7}(3),  966--977 (jun 2012)

\bibitem{Fierrez-Aguilar2004}
Fierrez-Aguilar, J., Alonso-Hermira, N., Moreno-Marquez, G., Ortega-Garcia, J.:
  An off-line signature verification system based on fusion of local and global
  information. In: Biometric Authentication, pp. 295--306. Springer (2004)

\bibitem{Fierrez-Aguilar2004TargetVerification}
Fierrez-Aguilar, J., Ortega-Garcia, J., Gonzalez-Rodriguez, J.: Target
  dependent score normalization techniques and their application to signature
  verification. IEEE Trans. on Systems, Man, and Cybernetics, Part C
  \textbf{35}(3),  418--425 (2004)

\bibitem{Fotak2011}
Fotak, T., Baca, M., Koruga, P.: {Handwritten signature identification using
  basic concepts of graph theory}. WSEAS Trans. on Signal Processing
  \textbf{7}(4),  145--157 (2011)

\bibitem{Gilperez2008Off-lineFeatures}
Gilperez, A., Alonso-Fernandez, F., Pecharroman, S., Fierrez, J.,
  Ortega-Garcia, J.: Off-line signature verification using contour features.
  In: {Proc. 11th Int. Conf. on Front. in Handwriting Rec.} pp.~1--6 (2008)

\bibitem{Hafemann2017}
Hafemann, L.G., Sabourin, R., Oliveira, L.S.: Learning features for offline
  handwritten signature verification using deep convolutional neural networks.
  Pattern Recognition  \textbf{70},  163--176 (2017)

\bibitem{Hafemann2017review}
Hafemann, L.G., Sabourin, R., Oliveira, L.S.: Offline handwritten signature
  verification - literature review. In: Proc of Int. Conf. on Image Processing
  Theory, Tools and Applications (IPTA). pp.~1--8 (Nov 2017)

\bibitem{he2016deep}
He, K., Zhang, X., Ren, S., Sun, J.: Deep residual learning for image
  recognition. In: Proc of Conf. on Computer Vision and Pattern Recognition.
  pp. 770--778 (2016)

\bibitem{hoffer2015}
Hoffer, E., Ailon, N.: Deep metric learning using triplet network. In:
  International Workshop on Similarity-Based Pattern Recognition. pp. 84--92.
  Springer (2015)

\bibitem{Impedovo2008}
Impedovo, D., Pirlo, G.: {Automatic signature verification: The state of the
  art}. IEEE Trans. on Systems, Man, and Cybernetics, Part C  \textbf{38}(5),
  609--635 (2008)

\bibitem{maergner2017icdar}
Maergner, P., Riesen, K., Ingold, R., Fischer, A.: A structural approach to
  offline signature verification using graph edit distance. In: Proc. of
  International Conference on Document Analysis and Recognition. pp.
  1216--1222. IEEE (2017)

\bibitem{Malik2012a}
Malik, M.I., Liwicki, M.: {From Terminology to Evaluation: Performance
  Assessment of Automatic Signature Verification Systems}. Proc. of Int.
  Conference on Frontiers in Handwriting Recognition pp. 613--618 (2012)

\bibitem{Ortega-Garcia2003}
Ortega-Garcia, J., Fierrez-Aguilar, J., Simon, D., Gonzalez, J., Faundez-Zanuy,
  M., Espinosa, V., Satue, A., Hernaez, I., Igarza, J.J., Vivaracho, C.,
  Escudero, D., Moro, Q.I.: {MCYT baseline corpus: a bimodal biometric
  database}. IEEE Proceedings-Vision, Image and Signal Processing
  \textbf{150}(6),  395--401 (2003)

\bibitem{Plamondon1989}
Plamondon, R., Lorette, G.: {Automatic signature verification and writer
  identification - the state of the art}. Pattern Recognition  \textbf{22}(2),
  107--131 (1989)

\bibitem{Riesen2009}
Riesen, K., Bunke, H.: {Approximate graph edit distance computation by means of
  bipartite graph matching}. Image and Vision Computing  \textbf{27}(7),
  950--959 (6 2009)

\bibitem{Riesen2014}
Riesen, K., Fischer, A., Bunke, H.: {Computing upper and lower bounds of graph
  edit distance in cubic time}. Int. Workshop on Artificial Neural Networks in
  Pattern Recognition  \textbf{8774},  129--140 (2014)

\bibitem{sabourin94structural}
Sabourin, R., Plamondon, R., Beaumier, L.: Structural interpretation of
  handwritten signature images. Int. Journal of Pattern Recognition and
  Artificial Intelligence  \textbf{8}(3),  709--748 (1994)

\bibitem{Yilmaz2011OfflineFeatures}
Yilmaz, M.B., Yanikoglu, B., Tirkaz, C., Kholmatov, A.: Offline signature
  verification using classifier combination of {HOG} and {LBP} features. In:
  {Proc. Int. Joint Conference on Biometrics}. pp.~1--7 (2011)

\bibitem{zagoruyko2015}
Zagoruyko, S., Komodakis, N.: Learning to compare image patches via
  convolutional neural networks. In: Proc of IEEE Conf. on Computer Vision and
  Pattern Recognition. pp. 4353--4361 (2015)

\end{thebibliography}

\end{document}